\documentclass[letterpaper, 10 pt, conference]{ieeeconf}  

\usepackage{caption}
\usepackage{subcaption}
\usepackage{algpseudocode}
\usepackage{algorithm}
\usepackage{graphicx}
\usepackage{amsmath}
\usepackage{multirow}
\usepackage{tabularx}
\usepackage{svg}
\usepackage{bbm}
\svgpath{{img/}}

\IEEEoverridecommandlockouts                              

\title{\LARGE \bf Optimization and Evaluation of Multi Robot Surface Inspection Through Particle Swarm Optimization}

\author{Darren Chiu$^{1}$, Radhika Nagpal$^{2}$, 
and Bahar Haghighat$^{3}$ 
\thanks{This work was supported by a Swiss National Science Foundation (SNSF) postdoctoral fellowship award P400P2\_191116 and an Office of Naval Research (ONR) grant N00014-22-1-2222
.}
\thanks{$^{1}$Darren Chiu is with University of Southern California, Los Angeles, CA 90089, USA
        {\tt\small chiudarr@usc.edu}}%
\thanks{$^{2}$Radhika Nagpal is with Princeton University, Princeton, NJ 08544, USA
        {\tt\small rn1627@princeton.edu}}%
\thanks{$^{3}$Bahar Haghighat is with University of Groningen, Groningen, 9747 AG, Netherlands
        {\tt\small bahar.haghighat@rug.nl}}%
}

\begin{document}

\maketitle
\thispagestyle{empty}
\pagestyle{empty}

\def\arraystretch{1.15}
\begin{abstract}

Robot swarms can be tasked with a variety of automated sensing and inspection applications in aerial, aquatic, and surface environments. In this paper, we study a simplified two-outcome surface inspection task. We task a group of robots to inspect and collectively classify a 2D surface section based on a binary pattern projected on the surface. 
We use a decentralized Bayesian decision-making algorithm and deploy a swarm of miniature 3-cm sized wheeled robots to inspect randomized black and white tiles of $1m\times 1m$. We first describe the model parameters that characterize our simulated environment, the robot swarm, and the inspection algorithm. We then employ a noise-resistant heuristic optimization scheme based on the Particle Swarm Optimization (PSO) using a fitness evaluation that combines decision accuracy and decision time. We use our fitness measure definition to asses the optimized parameters through 100 randomized simulations that vary surface pattern and initial robot poses. The optimized algorithm parameters show up to a 55\% improvement in median of fitness evaluations against an empirically chosen parameter set.

\end{abstract}
\setlength{\belowcaptionskip}{-3pt}

\section{INTRODUCTION}
\label{sec:intro}
Automated robotic inspection can serve many applications such as maintenance of bridges, wind turbines, oil and gas pipelines, and aerospace infrastructure \cite{bualat2008autonomous,carbone_swarm_2018,carrillo-zapata_mutual_2020,liu2022review}. In many of these instances, the inspection task takes the form of a binary classification problem. The classification goal is to determine the state of an inspected surface area as ``desirable'' or ``undesirable'' based on spatially-distributed surface features. A real-world example application of this scenario within the aerospace industry is inspection of tiled Thermal Protection System (TPS) surfaces of space shuttle hulls. For each tile, the goal is to conclude if it needs to be replaced based on an assessment of its surface defects. We propose deploying robot swarms to these types of tasks. Multiple benefits can be expected from deploying swarms of mobile robots. Compared to single-robot systems, swarms are resilient to failure of individuals. As for fixed sensor networks, robot swarms provide dynamic and flexible coverage performances \cite{bayat_environmental_2017}. In large-scale swarms, low-cost operations necessitates minimizing the cost and complexity of individual robots. This promotes the cause of swarms of miniaturized robots that employ computationally inexpensive algorithms.

The two-outcome swarm robotic surface inspection task that we consider here requires two main enabling mechanisms for the decision making: (i) a mechanism for the swarm to share and integrate observations made by individual robots, i.e., a sensor fusion mechanism, and (ii) a consensus mechanism to form a final classification decision out of the individual decisions, i.e., a decision fusion mechanism. A variety of distributed decision-making algorithms including Bayesian \cite{ebert2018multi,valentini2016collective2,valentini2016collective,valentini2015efficient}. 
 and bio-inspired \cite{alanyali2004distributed,bandyopadhyay2014distributed,zhao2007distributed,haghighat2022approach} algorithms have been employed in studies involving robot swarms. Compared to bio-inspired algorithms, Bayesian algorithms provide a statistically grounded approach for integrating the observations and decisions of the individual robots. 
In a previous contribution, a Bayesian approach was demonstrated in a swarm of abstract agents tasked with classifying a monochrome environment \cite{ebert2020bayes}. 
We build upon that work in three ways. First, we extend the algorithm presented in \cite{ebert2020bayes} by introducing a hysteresis parameter that defines a minimum observation criteria around the probability threshold before a robot updates its decision. Additionally, we adapt the algorithm for physically embodied robots by including proximity sensing and collision avoidance and present experiments using the physics-based Webots robotic simulator. Finally, we use a Particle Swarm Optimization (PSO) scheme to develop an automated optimization framework that leverages our Webots simulation world to heuristically find a set of optimized algorithm parameters for improved decision-making time and accuracy.   

This work strives to set a step towards developing algorithmic and hardware tools that support sensing tasks by robot swarms. Our insight is that (i) swarms of small-scale robots have the potential to deliver a variety of real world sensing and inspection applications, and that (ii) the statistically grounded Bayesian decision-making framework has the flexibility to be extended and applied to a broad class of spatially-distributed feature classification tasks by swarms.

\section{Problem Definition}
\label{sec:prob_def}
We define the problem that we set out to undertake as the following. A group of $N$ robots must complete a binary classification task based on perceiving a spatially distributed feature spread over a bounded 2D surface section. A black and white binary pattern representing the spatially distributed feature is projected on the surface. The pattern's fill ratio determines the proportion of the white-colored area in the overall pattern surface area. The robots each individually inspect the surface, share their information with the rest of the swarm, and collectively determine whether the surface is covered with a majority white or a majority black pattern.

The class of real-world inspection problems that underlies our abstract problem definition here is characterized by three main features: (i) the need to inspect a bounded surface environment, (ii) the need to deliver a binary classification decision for the inspected environment, and that (iii) the feature that informs the classification decision is spatially distributed in the environment where the robots operate. 
\section{Simulation and Optimization Framework}
\label{sec:sim_frame}
\begin{figure}[t]
    \centering
    \begin{subfigure}[t]{0.24\columnwidth}
        \centering
        \includegraphics[width = 1.0\columnwidth]{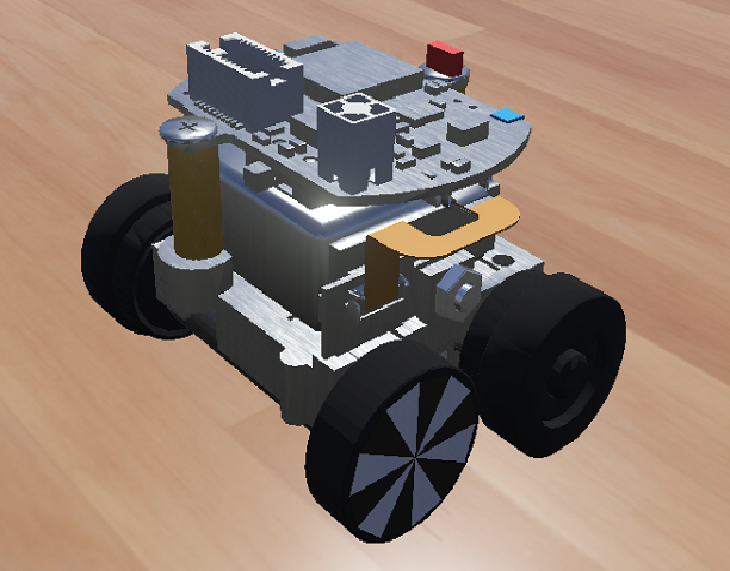}
        \caption{\scriptsize Simulation, side}
    \end{subfigure}
    \centering
    \begin{subfigure}[t]{0.24\columnwidth}
        \centering
        \includegraphics[width = 1.0\columnwidth]{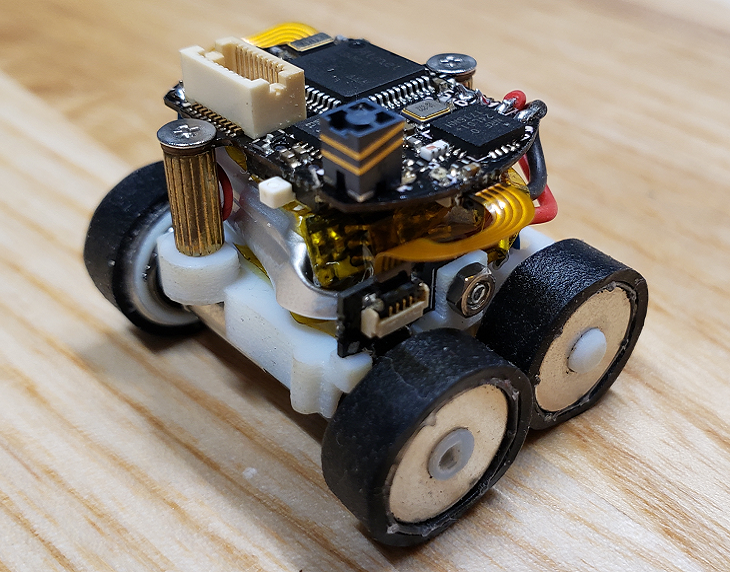}
        \caption{\scriptsize Real robot, side}
    \end{subfigure}
    \centering
    \begin{subfigure}[t]{0.24\columnwidth}
        \centering
        \includegraphics[width = 1.0\columnwidth]{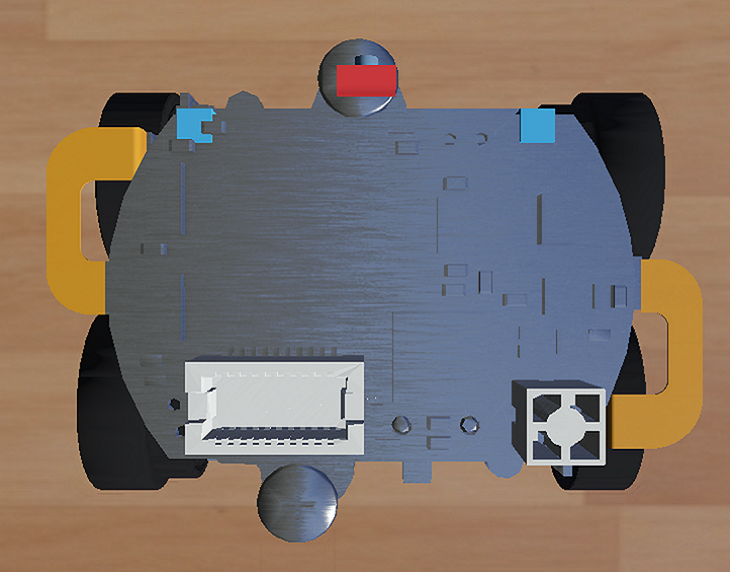}
        \caption{\scriptsize Simulation, top}
    \end{subfigure}
    \centering
    \begin{subfigure}[t]{0.24\columnwidth}
        \centering
        \includegraphics[width = 1.0\columnwidth]{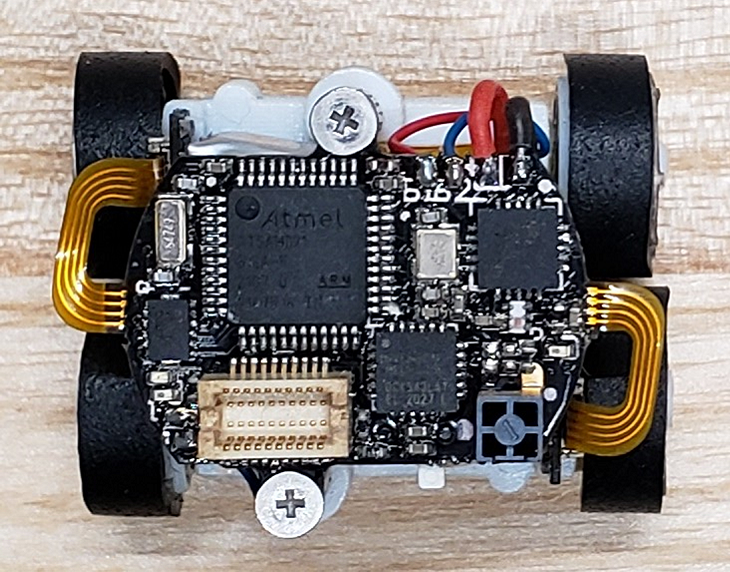}
        \caption{\scriptsize Real robot, top}
    \end{subfigure}
    \caption{We use a realistic model of the 4-wheeled Rovable robot in our simulation experiments \cite{dementyev_rovables_2016}. The real Rovable robot (b,d) and its simulation model created in Webots (a,c) have similar physical properties. The robots adhere to ferromagnetic surfaces using their magnetic pincher-wheels. For scale, each wheel is 12mm in diameter. 
    }
    \label{img:sim_rovable_model}
\end{figure}

\begin{figure}[t]
    \centering
    \begin{subfigure}[t]{0.24\columnwidth}
        \centering
        \includegraphics[width = 1.0\columnwidth]{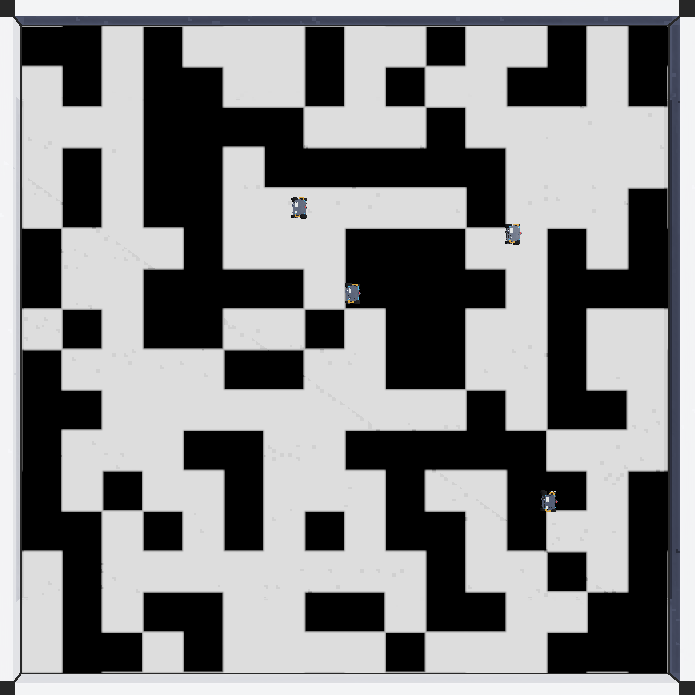}
        \caption{\scriptsize $f = 0.55$}
    \end{subfigure}
    \centering
    \begin{subfigure}[t]{0.24\columnwidth}
        \centering
        \includegraphics[width = 1.0\columnwidth]{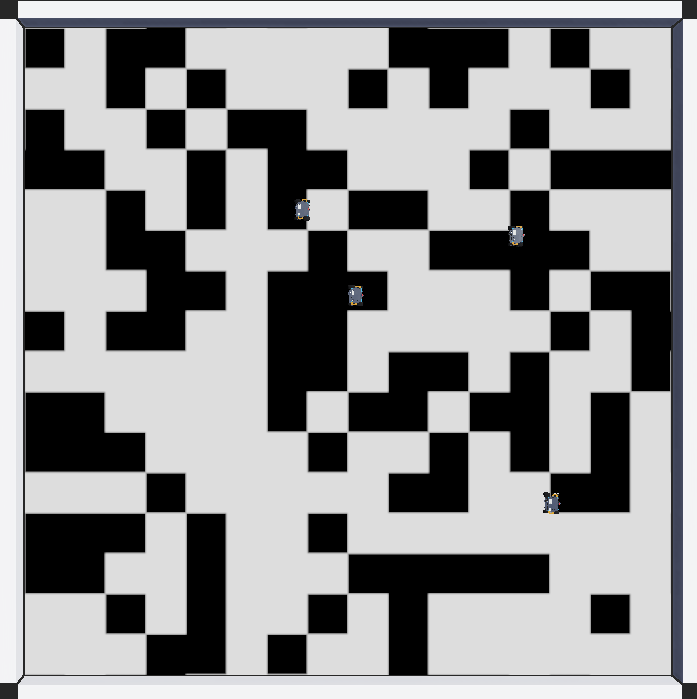}
        \caption{\scriptsize $f = 0.6$}
    \end{subfigure}
    \centering
    \begin{subfigure}[t]{0.24\columnwidth}
        \centering
        \includegraphics[width = 1.0\columnwidth]{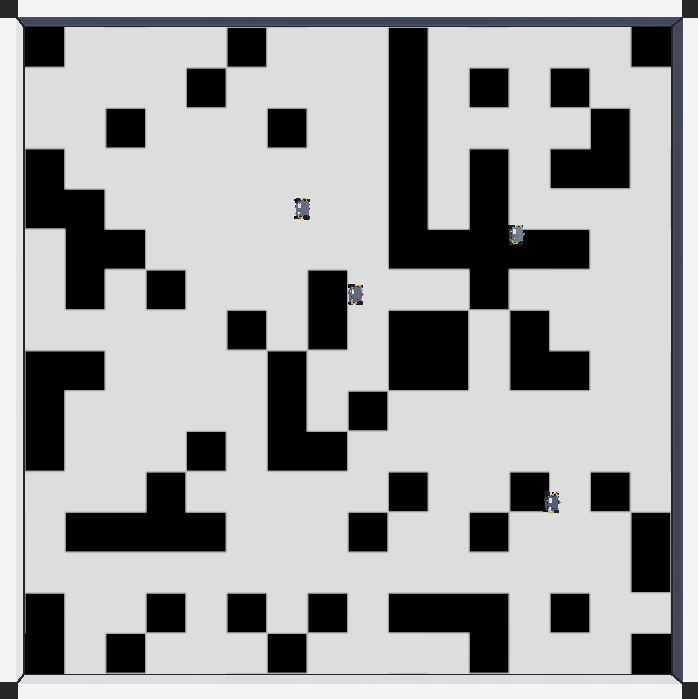}
        \caption{\scriptsize $f = 0.7$}
    \end{subfigure}
    \centering
    \begin{subfigure}[t]{0.24\columnwidth}
        \centering
        \includegraphics[width = 1.0\columnwidth]{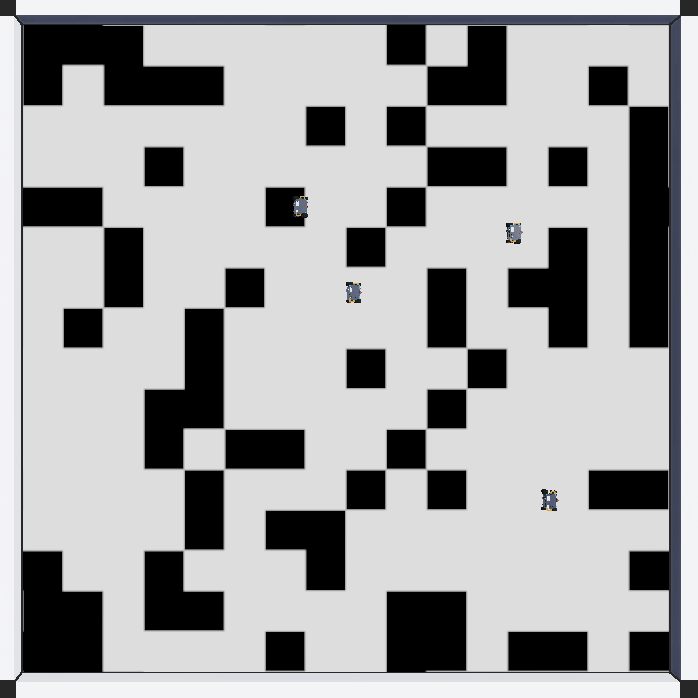}
        \caption{\scriptsize $f = 0.8$}
    \end{subfigure}
    \caption{Within the Webots robotic simulator, we simulate a bounded 2D surface section of size $1m \times 1m$ with a monochromatic randomized pattern. The proportion of white in the pattern, denoted as $f$, is the environment fill ratio. The robots have to classify the environment as mostly white ($f > 0.5$) or mostly black ($f < 0.5$). Environments with fill ratios close to $f=0.5$ are hardest to classify. There are a total of 256 squares in the arena, 16 squares along each side.
    }
    \label{img:sim_env}
\end{figure}

Our simulation framework serves as the virtual environment in which we study the operation of our inspecting robot swarm. We use the Webots robotic simulator \cite{michel_webotstm_2004}. Within Webots, we have three main components: (i) a realistic robot model of a 3-cm sized 4-wheeled robot that observes a target surface section, (ii) the target surfaces that the robots traverse to inspect with a black and white pattern projected onto them, and (iii) a (supervisor controller) script that collects robot positions, beliefs, decisions, and observations. The robot model, shown in Figure \ref{img:sim_rovable_model}, is based on the real Rovable robot \cite{dementyev_rovables_2016}. Originally designed as a mobile wearable robot, Rovables are capable of wireless communication and low-power localization using their wheel encoders and on-board IMUs for inertial-based navigation \cite{dementyev_rovables_2016}. 
Within Webots, the Rovable proto file captures the physical properties of the real robot, such as mass, center of mass, and surface contact properties. Figure \ref{img:sim_env} shows our overall Webots simulation world with four Rovable robots on four different patterned surfaces of size $1m \times 1m$ with different fill ratios.

Our optimization framework has two main components: (i) our Webots simulation framework described above and (ii) a PSO-based optimization scheme where each particle in the PSO swarm corresponds to an instance of our simulated world with a specific set of algorithm parameters. The optimization goal is to find a set of algorithm parameters that enable the robots to classify the environment with speed, accuracy, and consistency. Pugh et al. showed that PSO could outperform Genetic Algorithms on benchmark functions and in certain scenarios of limited-time learning with presence of noise \cite{Pugh2009a,Pugh2005}. This motivates the choice of PSO here since stochasticity is inherent to the approach that the robots employ. 
Multiple computationally efficient noise-resistant versions of PSO have been proposed \cite{di2014analysis,di2015distributed}. We simply evaluate each particle multiple times and consider a combination of average and standard deviation of observed performances.

The velocity of particle $i$ in dimension $j$ is determined by three components: (i) the particle's velocity at the previous iteration weighted by an inertia coefficient $w$, (ii) a randomized attraction to the particle's own personally best visited location over the previous iterations $x^*_{i,j}$ weighted by $w_p$, and (iii) a randomized attraction to the particle's neighborhood's best visited location over the previous iterations $x^*_{i',j}$ weighted by $w_n$~(Equation~\ref{eq:pso_vel}). $r_1$ and $r_2$ are random numbers drawn from a uniform distribution between 0 and 1. Attractions are determined through a fitness function we define that rewards fast and consistent decisions. Particle positions are then set at each time step using the updated velocity (Equation~\ref{eq:pso_pos}). 

\begin{subequations}
\begin{equation}
\label{eq:pso_vel}
v_{i,j}:=w \cdot v_{i,j}+ w_p \cdot r_1 \cdot (x^*_{i,j}-x_{i,j}) + w_n \cdot r_2 \cdot (x^*_{i',j}-x_{i,j})
\end{equation} \vspace{-0.5cm}
\begin{equation}
\label{eq:pso_pos}
x_{i,j}:=x_{i,j} + v_{i,j}
\end{equation}
\end{subequations}

The total execution time for the PSO optimization process depends on four factors: (i) population size ($N_p$), (ii) individual candidate evaluation time ($t_e$), (iii) number of iterations of the algorithm ($N_i$), and (iv) number of re-evaluations of each candidate solution's personal best within the same iteration ($N_{re}$). We evaluate all instances of the particles in the PSO swarm in parallel at each iteration, as a result the total time we get for the optimization procedure is as below:
\begin{equation}
\label{eq:tot_time}
t_{total}=t_e\cdot N_i
\end{equation}
The individual candidate evaluation time $t_e$ depends directly on the complexity of the simulation model utilized and its corresponding computational load.
\section{Inspection Algorithm}
\label{sec:algo}
The overall structure of our proposed algorithm is shown in Algorithm \ref{alg:alg_overview}. 
The algorithm enables simulated Rovable robots to inspect black and white 2D environments and classify the fill ratio as being above or below a predefined threshold level. Each robot in the swarm maintains a Bayesian model of the fill ratio that it updates using new self made observations and incoming observations from other robots. Each robot individually forms its decision about how to classify the fill ratio using a predefined credibility threshold and the posterior distribution. The robots make binary observations based on their location in the environment. We model the binary color observations $C \in \{0,1\}$ as drawings from a Bernoulli distribution. The fill ratio $f \in [0,1]$ is unknown to the robots and is modeled as a Beta distribution that is updated based on the incoming color observations. The Beta distribution is initialized with parameters $\alpha_0$ and $\beta_0$ as $Beta(\alpha = \alpha_0 , \beta = \beta_0)$, where $\alpha_0$ determines how regularizing the prior distribution is. 

\begin{subequations}
\begin{equation}
\label{eq:bernoulli}
C \sim Bernoulli(f)
\end{equation} \vspace{-0.5cm}
\begin{equation}
\label{eq:beta}
f \sim Beta(\alpha, \beta)
\end{equation}
\begin{equation}
\label{eq:post}
f \: | \: C \sim Beta(\alpha + C , \: \beta + (1-C))
\end{equation}
\end{subequations}

\begin{figure}[t]
    \centering
    \begin{subfigure}[t]{0.95\columnwidth}
        \centering
        \includegraphics[width = 0.95\columnwidth]{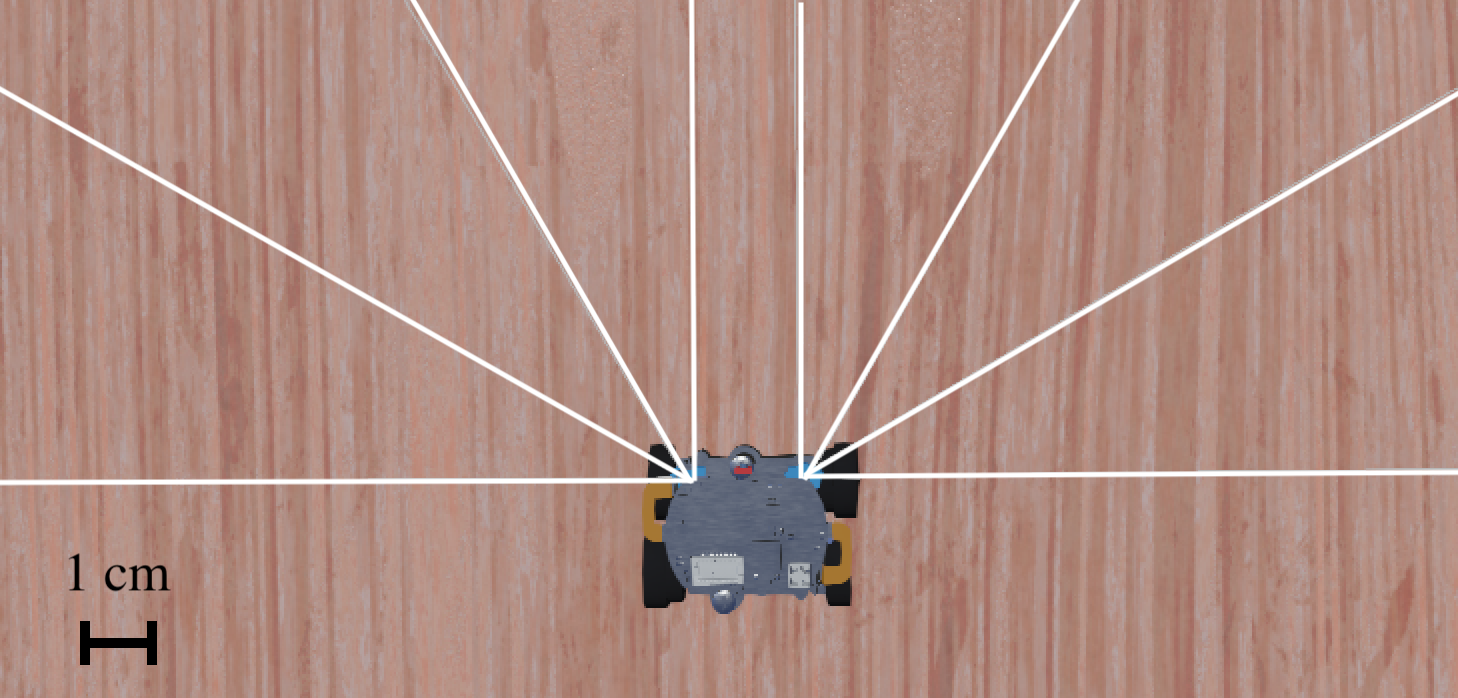}
    \end{subfigure}
    \caption{Simulated Rovable with distance sensors placed at 30 degree increments. A distance sensor is triggered by the parameter $d$ given in millimeters.}
    \label{img:rov_distance_sensor}
\end{figure}

The robots start inspecting the environment at randomized initial locations and orientations and explore the environment using a random walk while performing collision avoidance. Our random walk is implemented by drawing from a Gaussian distribution that is upper bounded by a parameter $s$, where $s$ defines the number of time steps the robot moves forward before taking a random turn. Each robot then turns by sampling a random angle drawn from a Gaussian distribution upper bounded by $\pi/2$. The collision avoidance is implemented using eight simulated time of flight sensors from the Webots library placed around the front of the robot (Figure \ref{img:rov_distance_sensor}). We introduce a parameter $d$ that defines the minimum sensed distance before a robot enters the collision avoidance state; $d$ applies to all eight sensors equally. 

\begin{algorithm}[t]
    \caption{Bayesian Inspection Algorithm}\label{alg:alg_overview}
    \hspace*{\algorithmicindent} \textbf{Input: $T, \alpha_o, \beta_o, \tau, f, r, h, p_c$} \\
    \hspace*{\algorithmicindent} \textbf{Output: $d_f \in\{0, 1\}$ }
    \begin{algorithmic}[1]
        \State $\alpha \gets \alpha_o$
        \Comment{Initialize alpha}
        \State $\beta \gets \beta_o$
        \Comment{Initialize beta}
        \State $d \gets -1$
        \Comment{Initialize incomplete decision flag}
        \While{$t \leq T$} 
            \State Perform Random Walk for $s$ Time Steps
            \If{$\tau \: divides \: t$}
                \State $Pause()$ 
                \Comment{Stop all movement for 40ms}
                \State $C \gets$ Observed Color
                \State $\alpha \gets \alpha + C$
                \State $\beta \gets \beta + (1-C)$
            \EndIf
            \State $C' \gets$ Message Color 
            \State $\alpha \gets \alpha + C'$
            \State $\beta \gets \beta + (1-C')$
            \State $p \gets Beta(\alpha, \beta, 0.5)$
            \If{$d_f \neq d_{f, t-1}$}
                \If{$p > p_c$ and $hysteresis$}
                    \State $d_f \gets 0$
                \ElsIf{$(1-p) > p_c$ and $hysteresis$} 
                    \State $d_f \gets 1$
                \EndIf
            \EndIf
            \If{$d_f \neq -1$ and $u^+$}
                \Comment{Broadcast decision}
                \State Broadcast $d_f$
            \Else 
                \Comment{Broadcast observation}
                \State Broadcast $C$
            \EndIf
            \State $t \gets t + 1$
        \EndWhile
    \end{algorithmic}
\end{algorithm} 

The robots make observations at regular intervals every $\tau$ simulation time steps from their location in the environment, update their posterior distribution as in Equation \ref{eq:post}, and broadcast their observation. The robots pause to sample the environment for 5 simulation steps (40$ms$ of real time). Upon receiving a radio message containing a new observation from an emitting robot, a receiving robot updates its posterior as it would using its own observations. This forms the decision fusion mechanism mentioned in Section \ref{sec:intro} in two configurations within the algorithm: with and without positive feedback. When positive feedback is false ($u^-$), emitting robots broadcast their most recent observation. When positive feedback is true ($u^+$), emitting robots instead broadcast their current decision. If no decision was made then the most recent observation is sent. After every posterior update, a robot also updates its classification decision using the predefined threshold $\theta$ and a credibility threshold $p_c$. The credibility threshold $p_c$ is defined as the probability mass of the posterior distribution that must lie on one side of the predefined threshold $\theta$ for the classification decision. In particular, if the posterior cumulative distribution $p$ at $f=\theta$ is greater than or equal to the credibility threshold $p_c$ then the classification decision may be set to 0 (majority black) or otherwise set to 1 (majority white). 

\begin{equation}
    \label{eqn:classification_decision}
    d_f=
    \begin{cases}
        1 & p(f=\theta) \geq p_c \\
        0 & p(f=\theta) < p_c 
    \end{cases}
\end{equation}

The hysteresis criterion must be met before setting the decision. This is determined using the hysteresis parameter denoted as $h$. The hysteresis parameter defines the minimum number of observations that must have been made after the posterior cumulative distribution, $p$, satisfies the credibility threshold, $p_c$. A non-zero hysteresis parameter enforces that only after $h$ observations have been made and $p$ continues to satisfy the condition described in Equation \ref{eqn:classification_decision}, the classification decision may be updated correspondingly. This is shown in Equation \ref{eqn:hysteresis_decision} where $o_i$ defines the number of observations after the condition in Equation \ref{eqn:classification_decision} is satisfied and $o(t)$ tracks the total number of observations made at time step $t$. Note that as new observations become available to a robot, its posterior cumulative distribution gets updated and the criteria defined in Equation \ref{eqn:classification_decision} may no longer be satisfied. In such a case, $o_i$ is reset to 0 and the hysteresis is broken.

\begin{equation}
    \label{eqn:hysteresis_decision}
    hysteresis=
    \begin{cases}
      1 & o(t) - o_i \geq h\\
      0 & o(t) - o_i < h
    \end{cases}
\end{equation}

\begin{figure}[t]
    \centering
    \begin{subfigure}[t]{\columnwidth}
        \centering
        \includegraphics[width = 1.0\columnwidth]{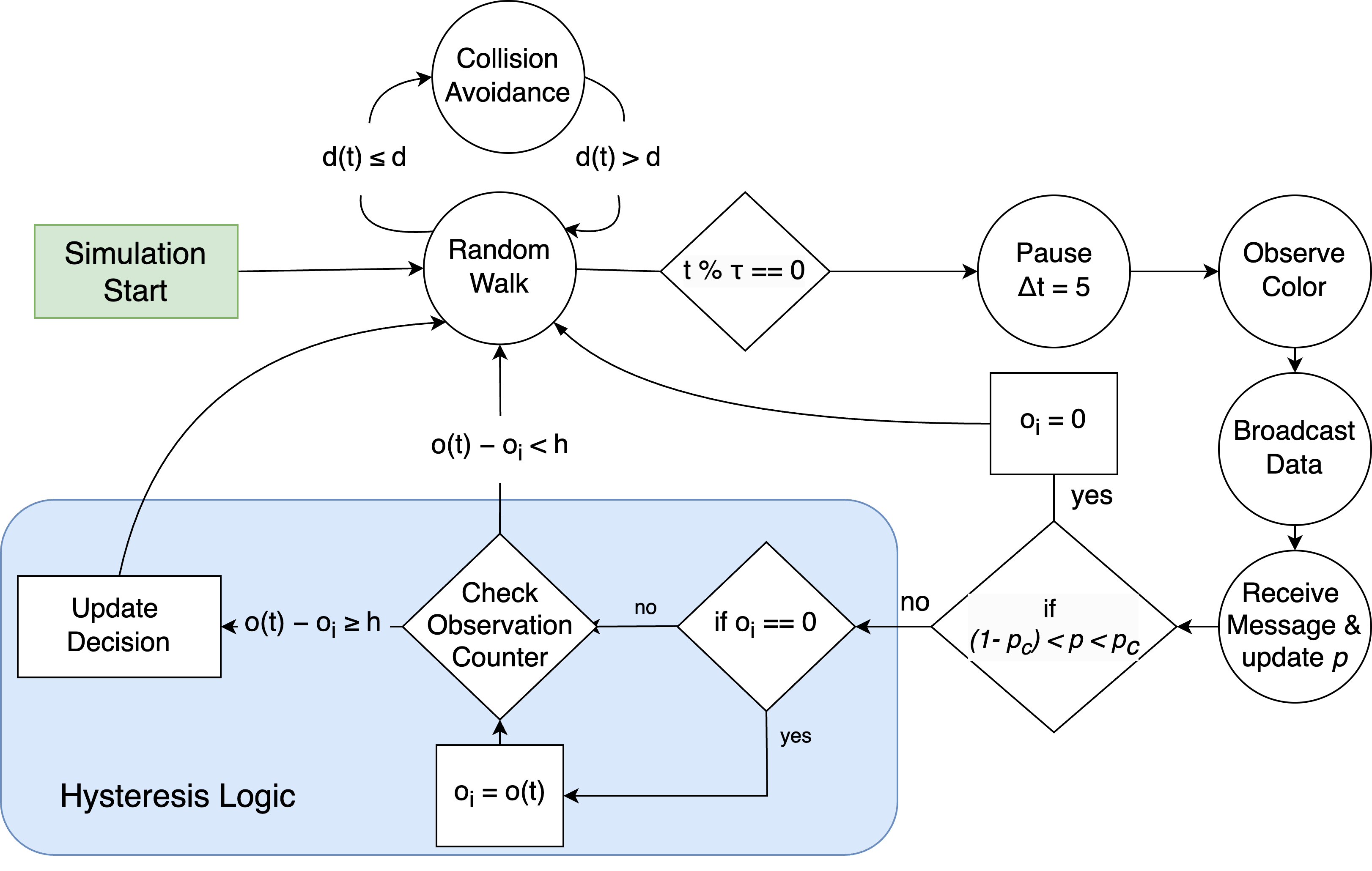}
    \end{subfigure}
    \caption{The finite state machine governing a robot's behavior. $d(t)$ indicates the measured distance from any sensor at time $t$. $o(t)$ tracks the total observations made up until simulation time step $t$. $o_i$ stores the initial number of observations once the credibility threshold is passed, for the purpose of calculating the hysteresis criterion. $h$ indicates the hysteresis parameter from Table \ref{table:parameter_bounds}. $p_c$ is the credibility threshold.
    }
    \label{img:bayes_fsm}
\end{figure}
We implement Algorithm \ref{alg:alg_overview} through the six-state finite state machine shown in Figure \ref{img:bayes_fsm}, where the states are denoted as circles, variable assignments as rectangles, and conditionals as diamonds. Each simulation step is $8ms$ long.

\section{Experiments}
\label{sec:experiments}
We used the Amazon Web Services (AWS) cloud platform for running the Webots simulations and the optimization process. Each simulation instance was launched on a 4-core CPU with 8GB of RAM and ran for a maximum of 60 minutes in simulation time, $T=3600s$. 
\begin{algorithm}[t]
    \caption{Fitness Evaluation}
    \begin{algorithmic}[1]
        \While{$t \leq T_{max}$}
        \Comment{Simulation loop}
            \If{New Decision for Robot $i$}
                \State{$n \gets n + 1$}
                \If{Decision is Correct}
                    \State{$f_i \gets f_i + t$}
                    \Comment{Add current time}
                \Else
                    \State{$f_i \gets f_i + T_{max}$}
                    \Comment{Add $T_{max} = 3600s$}
                \EndIf
            \EndIf
        \EndWhile
        \State{$f_i \gets \frac{f_i}{n}$}
        \If{Final decision is Incorrect}
            \State{$f_i \gets T_{max}$}
        \EndIf
        \State{return $fit = \sum^{N}_{i=0}f_i$}
        \Comment{$N$=number of robots}
    \end{algorithmic}
    \label{alg:evaluation}
\end{algorithm}

We use a swarm of four robots. The PSO parameters search space needs to be bounded before the PSO search is launched. In the following, we motivate the boundaries of our search space by considering physical features of the simulated arena. We bound the observation interval parameter, $\tau$, by deriving the number of simulation time steps needed for the robot to cross one colored square along one side, denoted as $t_s$. The random forward parameter, $s$, is bounded by deriving the time steps required to cross the arena along one side, denoted as $t_a$. The collision minimum distance parameter, $d$, was bounded by the range of the simulated sensors. The hysteresis parameter, $h$, was bounded by the number of squares in the arena, corresponding to the minimum number of attainable samples from the entire arena if a robot travelled one tile per sample. The lower and upper bounds of $\tau$ and $s$ were then expanded by an arbitrarily chosen factor of $5$ with the presumption that the optimal parameter would lie within these bounds. We assumed a perfect physics model of the robot that stops immediately on command with no skidding and traverses the arena at a constant speed (2.77 cm/s).

The PSO swarm is initialized randomly within the bounded search space, with the exception of one particle set to an empirically chosen location. We perform the optimizations with 15 particles, each evaluated 10 times for noise resistance, and proceed through 75 iterations.
For the empirical particle, the observation interval $\tau$ is set to the number of time steps needed to cross one square, $t_s=282$. The random forward variable $s$ is set to the number of time steps needed to cross two squares. We use an estimate of one robot body length to find the empirical collision avoidance distance $d=50$. Lastly, the hysteresis parameter $h$ is initialized at zero. This allows the optimization process to heuristically assess the utility of a non-zero hysteresis parameter. 
The final fitness assigned to a particle $p_i$ is then a weighted sum of the average ($\mu$) and the standard deviation ($\sigma$) across the noise evaluations as described in equation \ref{eqn:fitness}. $fit_{n}(p_i)$ denotes a fitness evaluation obtained through a single instance of particle $p_i$ following Algorithm \ref{alg:evaluation}. The parameter $\gamma$ is set to 1.1 empirically, giving a slightly higher significance to consistency compared with average performance. The optimization process favors lower fitness values as iterations progress, rewarding particles that return a low fitness value across multiple noise evaluations. 
\begin{equation}
    \label{eqn:fitness}
    fitness(p_i) = \mu(fit_{n}(p_i)) + \gamma\cdot \sigma(fit_{n}(p_i))
\end{equation}
\begin{table}[bh]
\caption{\label{table:parameter_bounds}$\alpha_0$: white observation prior parameter. $\beta_0$: black observation prior parameter. $\tau$: observation interval. $s$: random forward parameter. $d$: collision avoidance distance. $h$: hysteresis parameter. $n$: boundary multiplier, in our approach $n=5$. We derive $t_s=282$ and $t_a=4515$ using the robot speed (2.77 cm/s) and tile size (6.25cm) (see Section \ref{sec:experiments}).}
\begin{tabular}{|c|c|c|c|c|}
    \hline
    $\alpha_0=\beta_0 $ & $\tau$ & $s$ & $d$ & $h$\\ \hline
    \small $[0,0]$ & \small $[t_s/n,\:t_s\cdot n]$ & \small $[t_s/n,\:t_a \cdot n]$ & \small $[5,145]$ & \small $[0,\:128]$ \\
    \hline
    0 & 282 & 564  & 50 & 0 \\
    \hline
\end{tabular}
\end{table}

\section{Results}
\label{sec:results}
\begin{figure}[ht]
  \centering
  \begin{subfigure}[t]{1.0\columnwidth}
        \centering
        \includegraphics[width = 1.0\columnwidth]{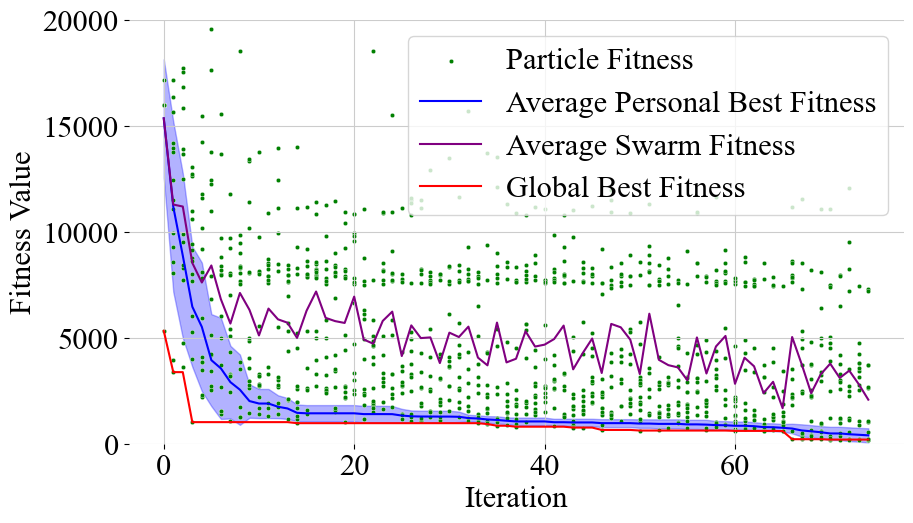}
        \vspace{-0.75cm}
        \caption{Fitness progression without positive feedback ($u^-$)}
      \vspace{0.18cm} 
  \end{subfigure}
  \begin{subfigure}[t]{1.0\columnwidth}
        \centering
        \includegraphics[width = 1.0\columnwidth]{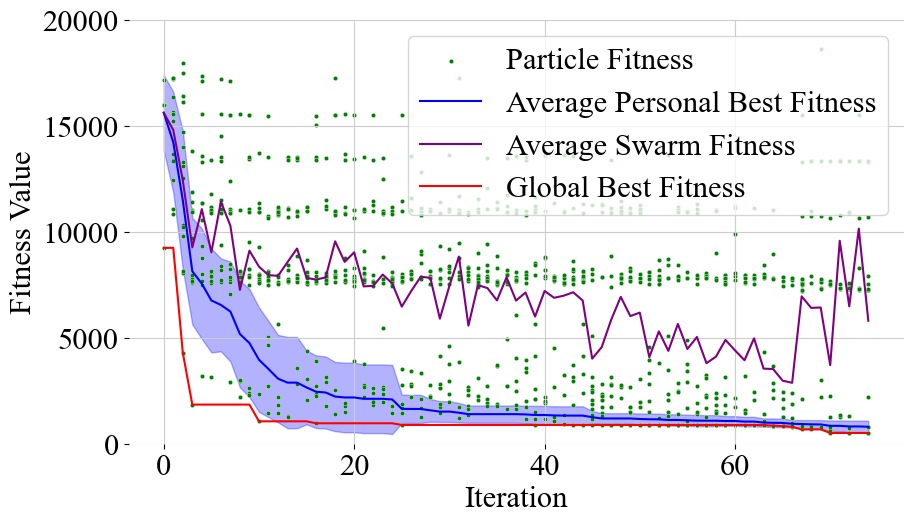}
        \vspace{-0.75cm}
        \caption{Fitness progression with positive feedback ($u^+$)}
    \end{subfigure}
    \caption{
    Optimization progression using a fitness evaluation that combines decision accuracy and speed (Equation \ref{eqn:fitness}), 15 particles, 10 noise evaluations, and for 75 iterations. We arrive at a best found fitness of 508.532 without positive feedback and 192.026 with positive feedback. Fitness values of the 15 particles (green) and the average swarm fitness (purple) are shown. 
    The average of personal bests (blue) can be seen to converge to the global best fitness value (red).
    }
    \label{fig:pso}
\end{figure}
\begin{table}[b]
    \caption{Empirical parameters are shown as $p$. Optimal parameters are shown as $p^*_{u^-}$ and $p^*_{u^+}$, for experiments without and with positive feedback, respectively.}
    \label{table:optimization_parameter_static}
    \begin{center} 
    \begin{tabular}{|c|c|c|c|c|c|}
    \hline
    Parameter Set & $\alpha$ & $\tau$ & $s$ & $d$ & $h$\\
    \hline
    $p$& 0 & 282 & 564 & 50 & 0 \\
    \hline
    $p^*_{u^-}$ & 0 & 56 & 178 & 29 & 17 \\
    \hline
    $p^*_{u^+}$ & 0 & 57 & 912 & 51 & 10 \\
    \hline
    \end{tabular}
    \end{center}
\end{table}

\begin{figure*}[t]
    \centering
    \begin{subfigure}[t]{2.0\columnwidth}
        \centering
        \includegraphics[width = 1.0\columnwidth]{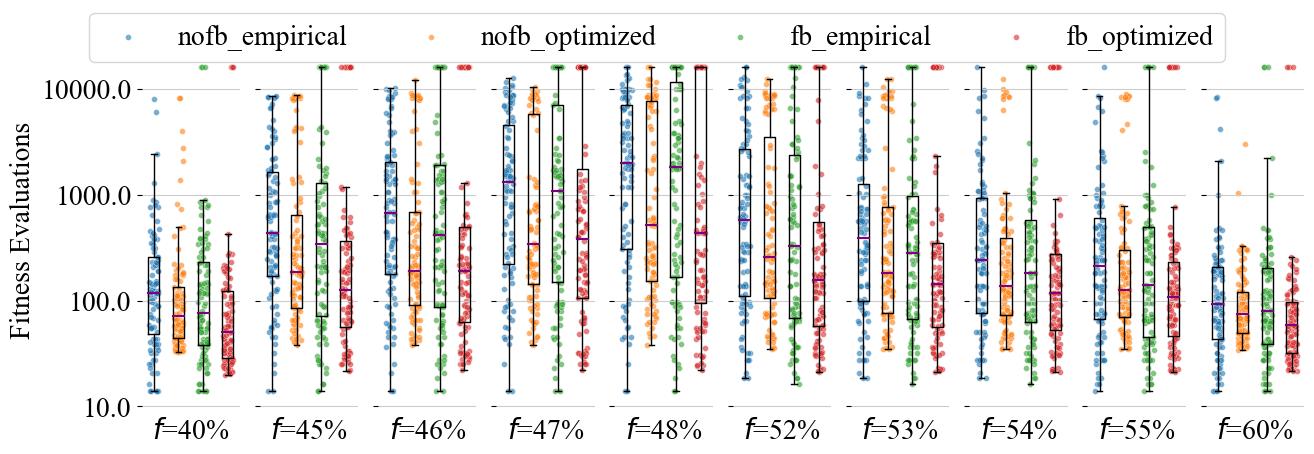}
    \end{subfigure}
    \caption{Distribution of fitness values obtained across 100 randomized simulations for each given fill ratio. We use the parameters described in Table \ref{table:optimization_parameter_static}. The optimized parameter set outperforms the empirical one in all cases, despite being specifically optimized for the case of $f=0.52$. We observe a $55.3\%$ reduction in the median evaluations through the optimization without positive feedback and a $51\%$ reduction in the median evaluations with positive feedback for $f=0.52$.}
    \label{img:fitness_static}
\end{figure*}
We ran our optimizations with 15 particles, 10 noise evaluations, and for 75 iterations. The results are shown in Figure \ref{fig:pso}. Through iterations, the average personal best fitness of the particles can be seen to converge downward towards the global best fitness. 
Since the simulation evaluation is noisy and particle exploration is stochastic, the individual particle fitness values, represented as the green dots, have large variations as iterations progress.
Table \ref{table:optimization_parameter_static} shows the optimal parameters. We examine the different optimal parameter values between the two cases of positive feedback states. In both optimal particles, the observation interval $\tau$ is brought down close to the lower bound. However the hysteresis parameter, $h$, was brought up despite the empirical particle having this parameter initialized to 0. This combination allows the robots to make faster observations (small $\tau$), yet remain safely elastic in their initial and subsequent decisions (non-zero $h$). The optimal value of random forward variable $s$ differs largely depending on the use of positive feedback. When positive feedback is not used, $s$ is set lower; each robot works to greedily gather data and shape individual decisions. On the contrary, robots spend more time exploring the arena when positive feedback is used, as indicated by a larger optimal value for $s$. In this case, individual robot decisions become heavily influential. A similar rationale applies to the smaller collision avoidance parameter $d$ obtained when no positive feedback is used; collision avoidance becomes less critical and robots greedily prioritize sampling.    

\begin{figure}
    \begin{subfigure}[t]{0.49\columnwidth}
        \centering
        \includegraphics[width = 1.0\columnwidth]{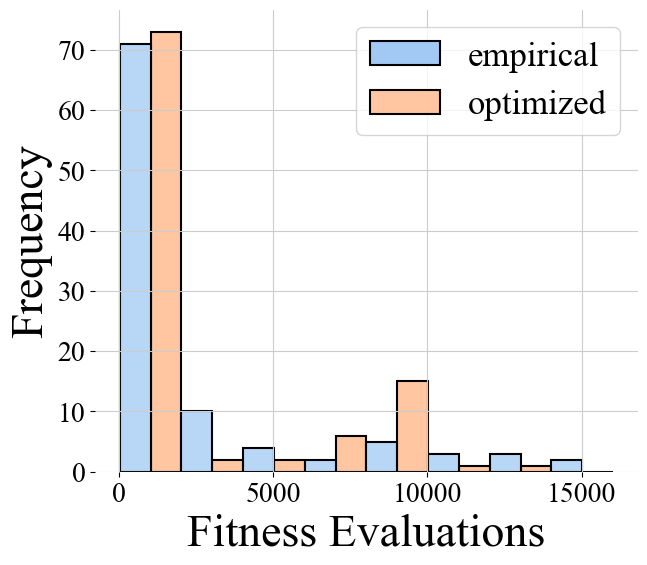}
        \caption{No positive feedback ($u^-$)}
    \end{subfigure}
    \centering
    \begin{subfigure}[t]{0.49\columnwidth}
        \centering
        \includegraphics[width = 1.0\columnwidth]{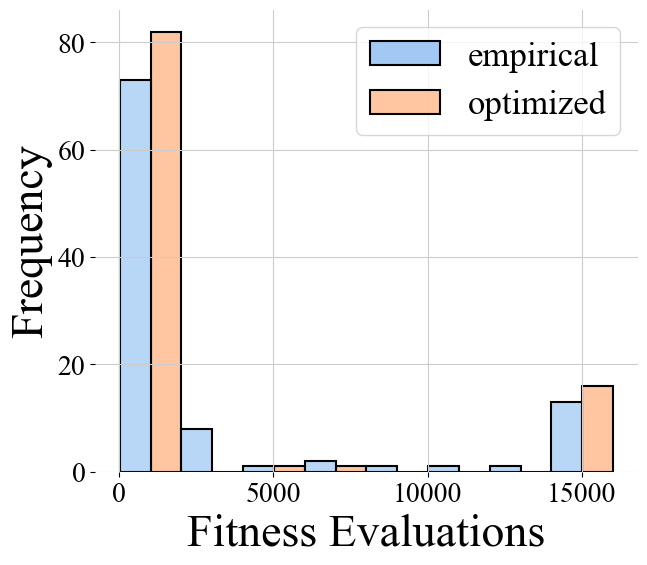}
        \caption{With positive feedback ($u^+$)}
    \end{subfigure}
    \caption{Distribution of fitness evaluations over 100 randomized simulations using a fill ratio of $f=0.52$. We observe that through optimization the distribution of fitness values shifts favorably towards lower values, in both cases with and without positive feedback being used.}
    \label{fig:eval_distribution}
\end{figure}

We evaluate the performance of the optimized particle against the empirical particle through 100 randomized experiments; varying the generated ($f=0.52$) pattern projected on the arena and initial robot poses. Figure \ref{img:fitness_static} shows the distribution of fitness values obtained across different fill ratios. Each dot represents a single particle evaluation calculated using Algorithm \ref{alg:evaluation}. The optimized parameters achieve a significantly lower median evaluation and more consistent performance compared to the empirical parameters.
\begin{figure}
        \begin{subfigure}[t]{0.49\columnwidth}  
        \centering
        \includegraphics[width = 1.0\columnwidth]{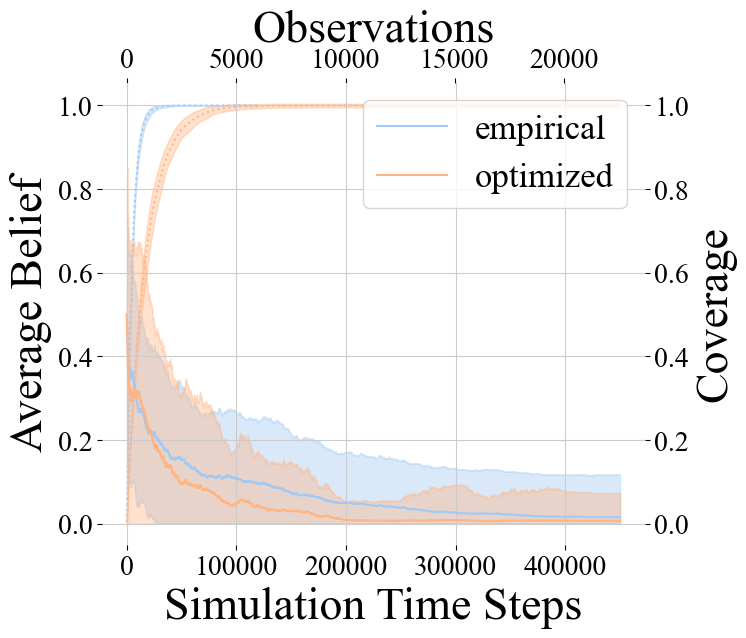}
        \caption{No positive feedback ($u^-$)}
    \end{subfigure}
            \centering
        \begin{subfigure}[t]{0.49\columnwidth}
        \centering
        \includegraphics[width = 1.0\columnwidth]{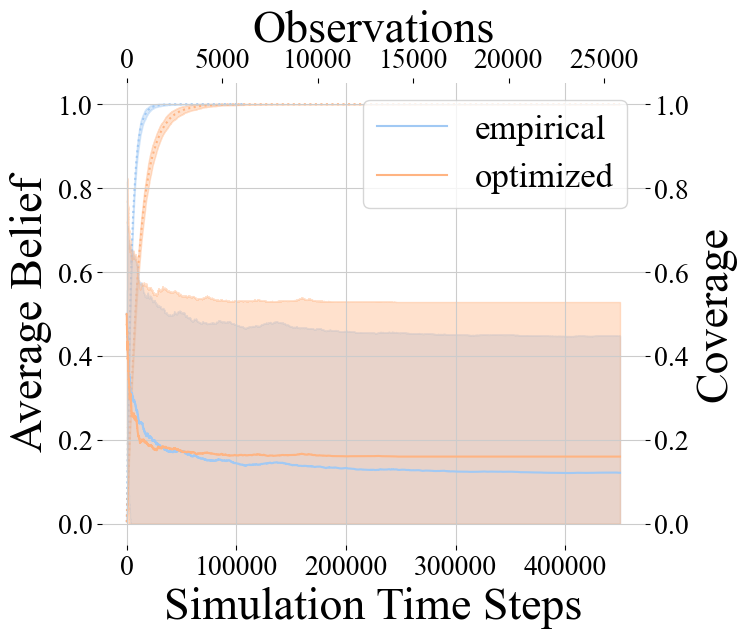}
        \caption{With positive feedback ($u^+$)}
    \end{subfigure}
    \caption{Average beliefs and coverage of the robots over 100 randomized simulation experiments. The simulations were randomized by rearranging the projected pattern (with the same fill ratio) and initial robot poses. The pattern fill ratio was fixed to $f=0.52$ in all cases. The optimized parameters lead the robots to converge faster towards the correct belief.}
    \label{fig:belief_distribution}
\end{figure}
The obtained fitness distributions for a fill ratio of $f=0.52$ across the 100 randomized experiments are shown in Figure \ref{fig:eval_distribution}. The distribution of fitness values shift towards the favorable side with and without positive feedback. Furthermore, the optimized cases show faster stabilization towards the correct beliefs (Figure \ref{fig:belief_distribution}). However, the experiments with positive feedback stabilize on average at a higher belief, exhibiting more incorrect classifications compared to those without positive feedback. This occurs when robots make quick decisions that spread across the group, leading to fast convergence, but towards the incorrect decision. Interestingly, the optimized parameters require the swarm to sample new tiles at a lower rate. Despite this, it can be seen in Figure \ref{fig:eval_distribution} that the robots are making faster correct classifications. We see that the non-zero hysteresis parameter ensures stability in the classification decision despite the apparent under-sampling, eventually benefiting the decision fusion mechanism through positive feedback.
In contrast, over-sampling without hysteresis ($h=0$) could quickly lead robots to misleading classification decisions. This is particularly detrimental when positive feedback is used, as the broadcasted incorrect decisions become adversarial.  

\section{Conclusion and Future Work}
\label{sec:conclusion}
In this paper, we presented a simulation and optimization framework for studying a two-outcome surface inspection task using a swarm of miniaturized wheeled robots that employ a decentralized Bayesian algorithm. We built upon a previously studied inspection algorithm by introducing a new hysteresis parameter that creates elasticity around robot decisions. 
We used the Webots robotic simulator to perform accurate physics-based simulations of a real robot swarm. Using a noise resistant variant of the Particle Swarm Optimization (PSO) method, we obtained a set of optimal parameters. 
Results from 100 randomized experiments revealed that the robot swarm employing optimized algorithm parameters was able to achieve a $55\%$ improvement in median fitness evaluations without positive feedback and a $51\%$ improvement with positive feedback compared to the empirically chosen parameters. Furthermore, the results revealed that a non-zero hysteresis parameter leads to improvements in the final decision accuracy. 
In future work, we plan to study two-outcome surface inspection tasks of higher complexity. In particular, we will (i) conduct simulations for inspection of geometrically complex 3D surface sections and consider dynamic environments, (iii) study commonly employed sensing modalities and signal processing methods such as camera feed or vibration sensing, and (vi) conduct real life experiments to validate the performance of the optimized parameter set obtained from simulation in reality.
\section*{ACKNOWLEDGMENT}

We wish to thank Dr. Olivier Michel and Yannick Goumaz from Cyberbotics Ltd. for supporting the AWS simulations and Dr. Ariel Ekblaw for making available the robot proto file for our simulations.



\bibliographystyle{IEEEtran}
\bibliography{./jb_ext}

\end{document}